\documentclass[lettersize,journal]{IEEEtran}
\usepackage{amsmath,amsfonts}
\usepackage{algorithmic}
\usepackage{algorithm}
\usepackage{array}

\usepackage[subrefformat=parens,labelformat=parens]{subcaption}
\usepackage[normalem]{ulem} 
\usepackage{textcomp}
\usepackage{stfloats}
\usepackage{url}
\usepackage{verbatim}
\usepackage{graphicx}
\usepackage{cite}
\usepackage{booktabs}  
\usepackage{multirow}  

\begin{document}

\title{Bitstream Action Recognition is Byte Modeling}
 

\author{
  Fangcheng~Li, 
  Chaoran~Huang,
  Tianyi~Liu,  
  Wenyang~Liu,
  Kejun~Wu,~\IEEEmembership{Senior Member,~IEEE},
  Qiong~Liu,~\IEEEmembership{Senior Member,~IEEE},
  You~Yang,~\IEEEmembership{Senior Member,~IEEE},
  and Zhengguo~Li,~\IEEEmembership{Fellow,~IEEE}
  \thanks{Fangcheng~Li, Chaoran~Huang, Kejun~Wu, Qiong~Liu, and You~Yang are with the School of Electronic Information and Communications, Huazhong University of Science and Technology, Wuhan 430074, China.}
  \thanks{Tianyi~Liu and Wenyang~Liu are with School of Electrical and Electronics Engineering, Nanyang Technological University, Singapore.}
  \thanks{Zhengguo~Li is with VI Department, Institute for Infocomm Research, Agency for Science, Technology and Research (A*STAR), Singapore.}  
  \thanks{This work was supported by the National Natural Science Foundation of China under Grant 62501246.}
  \thanks{Corresponding author: Kejun~Wu (kjwu@hust.edu.cn).}
  }

\markboth{Journal of \LaTeX\ Class Files,~Vol.~14, No.~8, August~2026}%
{Shell \MakeLowercase{\textit{et al.}}: A Sample Article Using IEEEtran.cls for IEEE Journals}

\maketitle

\begin{abstract}
Conventional action recognition typically relies on successful pixel decoding of the bitstream. However, bitstream corruption during storage or transmission may cause severe visual artifacts or even decoding failure, posing a significant challenge to reliable action recognition. Bitstream Action Recognition (BAR) aims to overcome the dependency on decoding and the vulnerability to corruption. In this paper, we propose a novel BAR framework, \uline{\textbf{B}}itstream \uline{\textbf{R}}ecognition via \uline{\textbf{A}}nchoring \uline{\textbf{C}}orrupted \uline{\textbf{E}}mbeddings (BRACE). BRACE is a dual-branch byte-modeling architecture that treats a corrupted bitstream and its intact counterpart as two byte realizations of the same action. This guides the generation of rich and stable representations for robustness to corruption through Intact-Anchored Representation Alignment (IARA). The intact representation serves as a stable anchor, and the corrupted one is aligned to it at the embedding and decision levels under Unreliable-Anchor Suppression (UAS), entirely in representation space and without repairing the bitstream. To address the scarcity of corrupted bitstreams in practice, we introduce the Real-world Bitstream Corruption Simulator (RBCS), a four-parameter simulator that reproduces bit-flip and byte-loss errors arising in transmission and storage. Building on RBCS, we construct the first large-scale BAR dataset (BAR-D), which comprises the BAR-Stanford40 and BAR-PPMI subsets and spans diverse corruption types and severity levels. Finally, we build a large benchmark on BAR-D involving 14 action recognition methods from the pixel, compressed, and bitstream domains. Extensive experiments demonstrate that BRACE has superior robustness to bitstream corruption than all comparison methods. Ablation studies further validate the effectiveness of the proposed RBCS augmentation and IARA.
\end{abstract}

\begin{IEEEkeywords}
Bitstream Action Recognition, Representation Alignment, Corruption Robustness.
\end{IEEEkeywords}

\section{Introduction}
\begin{figure}
  \centering
  \includegraphics[width=1\linewidth]{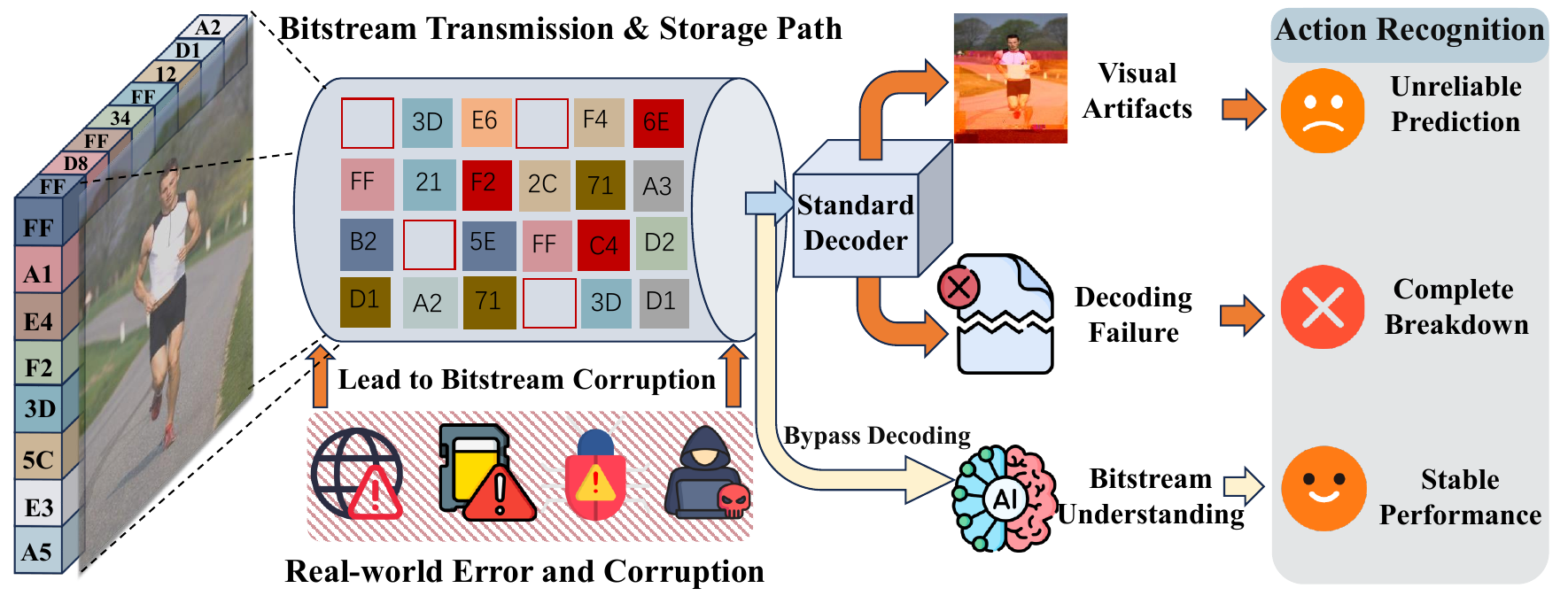}
  \caption{Real-world corruption of image bitstream during transmission and storage may cause severe visual artifacts or decoding failure, which can hardly be solved by standard image codecs and further affect image action recognition. Bitstream Action Recognition (BAR) enables the direct understanding of corrupted bitstream, solving the challenge of dependency on decoding and vulnerability to corruption.
  }\label{fig:Motivation}
\end{figure}

\IEEEPARstart{A}{CTION} recognition aims to identify the actions or behaviors presented in visual data~\cite{Sun2020HumanAR}. It plays an important role in a wide range of real-world applications, such as embodied artificial intelligence~\cite{11417250}, behavior analysis~\cite{10948348}, human object interaction~\cite{11249431}, and autonomous driving~\cite{11071878}. Although deep learning has greatly advanced action recognition, existing methods still rely on the successful decoding of complete bitstream into image pixels. Whether the bitstream is generated by traditional standard codecs or emerging end-to-end deep compression frameworks~\cite{wu2025end}, a standard decoder first converts the stored or transmitted bitstream into a visual image, from which the recognition model extracts discriminative cues such as human posture, object context, and scene layout for action classification. However, the reliability of this pipeline can be seriously affected when bitstreams are corrupted~\cite{wang1998error}. As shown in Fig.~\ref{fig:Motivation}, real-world errors may occur along the transmission and storage path, including unstable communication channels, physical storage damage, file system errors, and adversarial attack~\cite{382489}. These errors may flip or drop bytes and further damage the critical syntax information required by standard decoders, such as headers, markers, and coding tables~\cite{sencar2009identification}. As a result, the standard decoder produces severe artifacts or fails to recover valid pixels~\cite{liu2023restore,lin2026compressed}, which limit pixel-domain action recognition.

To handle unreliable visual inputs, existing studies mainly explored two directions, pixel-domain restoration and compressed-domain understanding. Pixel-domain restoration aims to recover degraded or incomplete content from decoded images before recognition~\cite{liu2023bitstream,liu2026promptsr}. However, it still requires the corrupted file to be decoded into a usable image, and therefore cannot work when the decoding process fails. Compressed-domain understanding attempts to use coding information inside compressed files, such as Discrete Cosine Transform (DCT) coefficients or other intermediate components~\cite{ehrlich2019deep}. However, it also requires the compressed data to be correctly parsed by standard decoders. Once severe bit-level corruption damages the key syntax of the encoded file, both decoded images and compressed-domain information may become unavailable. As a result, existing restoration-based and compressed-domain methods are insufficient for action recognition when bitstream corruption prevents reliable decoding or parsing.

Consequently, there is a need to move beyond decoder-dependent action recognition and explore direct modeling of raw image bitstreams. Recently, researchers have begun to learn directly from low-level encoded data, aiming to uncover the underlying visual semantics in binary bitstream~\cite{hortonbytes}. These studies show that image bitstreams are not merely files for storage and transmission, but can also serve as informative visual representations~\cite{10812851}. This progress provides a promising foundation for Bitstream Action Recognition (BAR), where actions are inferred directly from image bitstreams without reconstructing images. However, BAR remains challenging. Unlike pixel images, bitstreams do not present explicit spatial structures such as human regions, object layouts, or scene context. Unlike natural language, individual bytes also do not carry clear human-readable semantics. Instead, action-related cues are implicitly embedded in long, codec-dependent byte patterns. When random bitstream corruption occurs, these patterns may be disturbed or partially lost. Moreover, existing bitstream-domain studies focus on intact bitstreams, leaving corrupted bitstream modeling for action recognition largely unexplored.

To address this gap, we first introduce a Real-world Bitstream Corruption Simulator (RBCS). RBCS reproduces the bit-flip and byte-loss errors that occur in transmission and storage. The first large-scale BAR dataset BAR-D is constructed via RBCS.   
We then propose a novel BAR architecture, \uline{\textbf{B}}itstream \uline{\textbf{R}}ecognition via \uline{\textbf{A}}nchoring \uline{\textbf{C}}orrupted \uline{\textbf{E}}mbeddings  (BRACE) for BAR byte modeling. 
BRACE uses the intact counterpart of a corrupted bitstream as a reference and learns stable action representations through Intact-Anchored Representation Alignment (IARA). 
Our main contributions are summarized as follows.
\begin{itemize}
    \item To overcome the absence of BAR data, we introduce RBCS to generate corrupted bitstream. RBCS is a four-parameter simulator that injects bit-flip and byte-loss errors at controllable positions and rates in bitstream. By leveraging RBCS, we build the first large-scale dataset BAR-D. It contains 164{,}892 corrupted image bitstreams, comprising 4 severity levels of corruption, 3 corruption types of bit-flip, byte-loss, and their mixture, from the BAR-Stanford40 and BAR-PPMI subsets across 64 action classes.
    \item We propose BRACE, a dual-branch byte-modeling BAR architecture. BRACE treats a corrupted bitstream and its intact counterpart as two byte realizations of the same action. In the byte modeling of BRACE, the key component IARA pulls the corrupted representation to the intact anchor by embedding and decision alignments, while the Unreliable-Anchor Suppression (UAS) restricts this alignment to reliable anchors. In this way, BRACE enables action recognition from corrupted bitstreams by modeling their byte sequences.
    \item We build a large benchmark on BAR-D, providing a comprehensive comparison that spans the pixel, compressed, and bitstream domains with 14 action recognition methods. Across this benchmark, BRACE delivers superior corruption robustness over all comparison methods. Ablation studies further demonstrate the effectiveness of the proposed RBCS augmentation and IARA.
\end{itemize}

\section{Related Work}\label{sec:Relate}

\subsection{Pixel Domain Action Recognition}

Action recognition (AR) aims to identify the action present in a visual input. The task spans both video and still images. Video methods rely on temporal and motion cues across frames, while still-image AR must infer the action from a single frame. We focus on the still-image setting. Early work described each image with hand-crafted appearance features such as HOG and SIFT under a Bag-of-Visual-Words representation. Later methods added structured cues by modeling human pose configurations~\cite{bourdev2009Poselets} and the mutual context between people and the interacted objects. These designs relied on manually engineered features, which limited their capacity and generalization~\cite{guo2014survey}. The rise of convolutional neural networks (CNNs) shifted AR toward end-to-end representation learning~\cite{Krizhevsky2017AlexNet}. Region-based models reformulated the task as classification over human and object proposals~\cite{Gkioxari2015RCNN}, and deep residual backbones became standard feature extractors for transfer learning. Large-scale human-object interaction datasets such as HICO then accelerated data-driven study~\cite{Chao2015HICO}, and multi-branch detectors jointly localized people, objects, and their interactions~\cite{chen2025ask}. More recent work adopts Transformers for long-range dependencies~\cite{xie2025instructive} and leverages pre-trained vision-language models to improve the generalization of action recognition \cite{han2026training}. Across this progression, action recognition has been studied almost entirely on the decoded pixel image. Recognition from inputs other than a clean pixel image has drawn far less attention.

\subsection{Bitstream Domain Understanding}

Conventional visual understanding pipelines rely on decoding compressed images into pixels before semantic analysis. To reduce the cost of full pixel reconstruction, researchers have explored recognition beyond the pixel domain. Early compressed-domain methods extract coding cues from compressed images for retrieval and classification~\cite{jamil2019optimal}. Recent work further explores learning directly from encoded image data. Piau \textit{et al.} trained CNNs on entropy-coded images for image classification~\cite{piau2023learning}. Hill \textit{et al.} investigated image classification using compressed binary streams~\cite{hill2021transform}. These works show that compressed representations contain useful semantic cues. However, they still require correctly parsed coding structures, which may become unavailable when bitstream corruption damages key syntax information. Bitstream-domain understanding further reduces the reliance on decoded pixels and parsed compressed structures by learning directly from raw bytes. ByteFormer performs classification on common image file bytes with byte embeddings and a Transformer encoder~\cite{hortonbytes}. Beyond visual recognition, recent byte-level models also show the potential of raw-byte modeling for general digital data. bGPT learns unified representations from raw bytes across different data types~\cite{wu2024beyond}. MEGABYTE uses hierarchical patching to process million-length byte sequences~\cite{yu2023MEGABYTE}. MambaByte and MBLM further explore efficient token-free or hierarchical byte modeling for long sequences~\cite{wang2024MambaByte,egli2025multiscale}. These studies show that raw bitstreams contain learnable semantic patterns. However, they mainly focus on intact inputs and do not address recognition from corrupted image bitstreams.

\begin{figure*}[t]
  \centering
  \begin{subfigure}{0.47\textwidth}
    \centering
    \includegraphics[width=\textwidth]{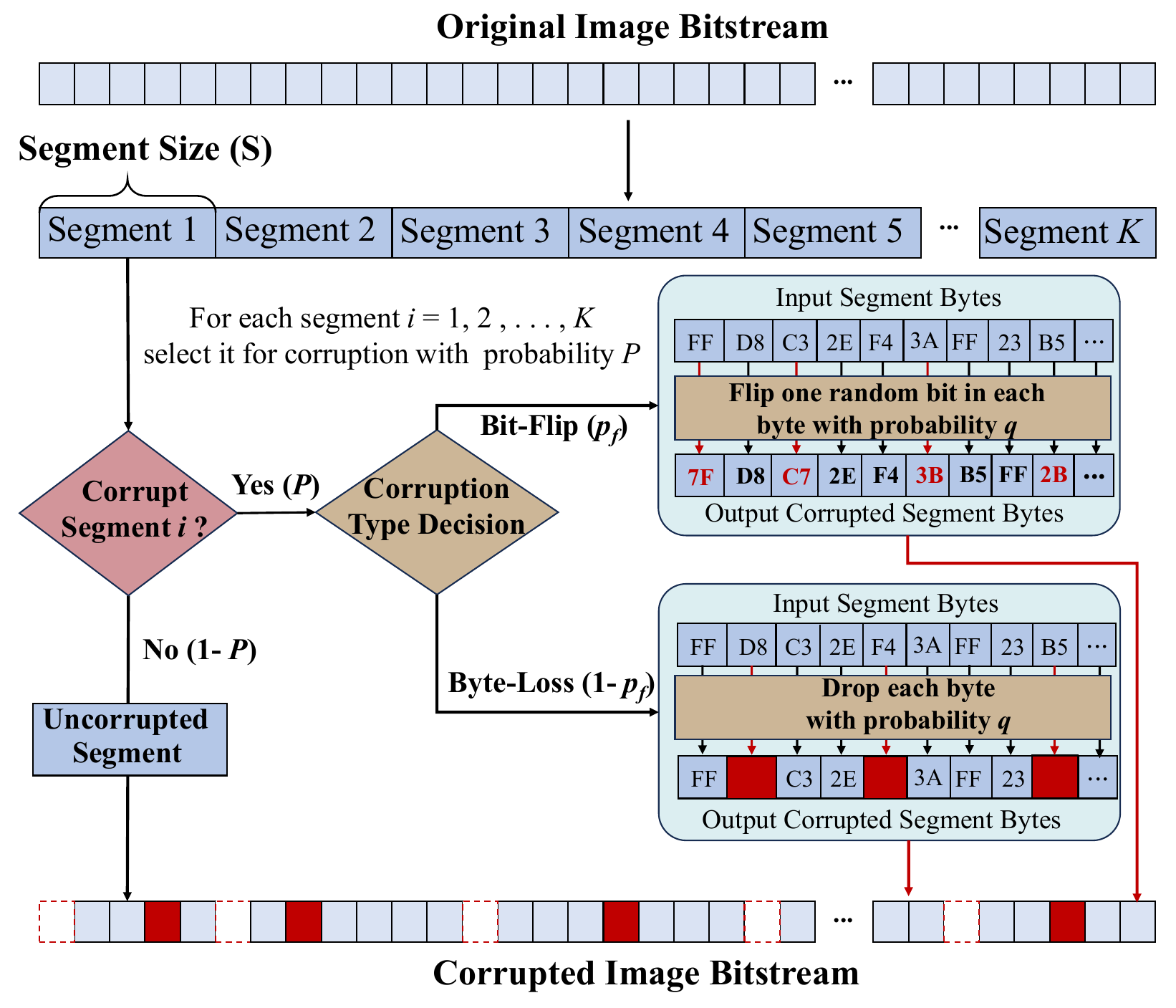}
    \caption{}
    \label{fig:RBCS_a}
  \end{subfigure}
  \hfil
  \begin{subfigure}{0.52\textwidth}
    \centering
    \includegraphics[width=\textwidth]{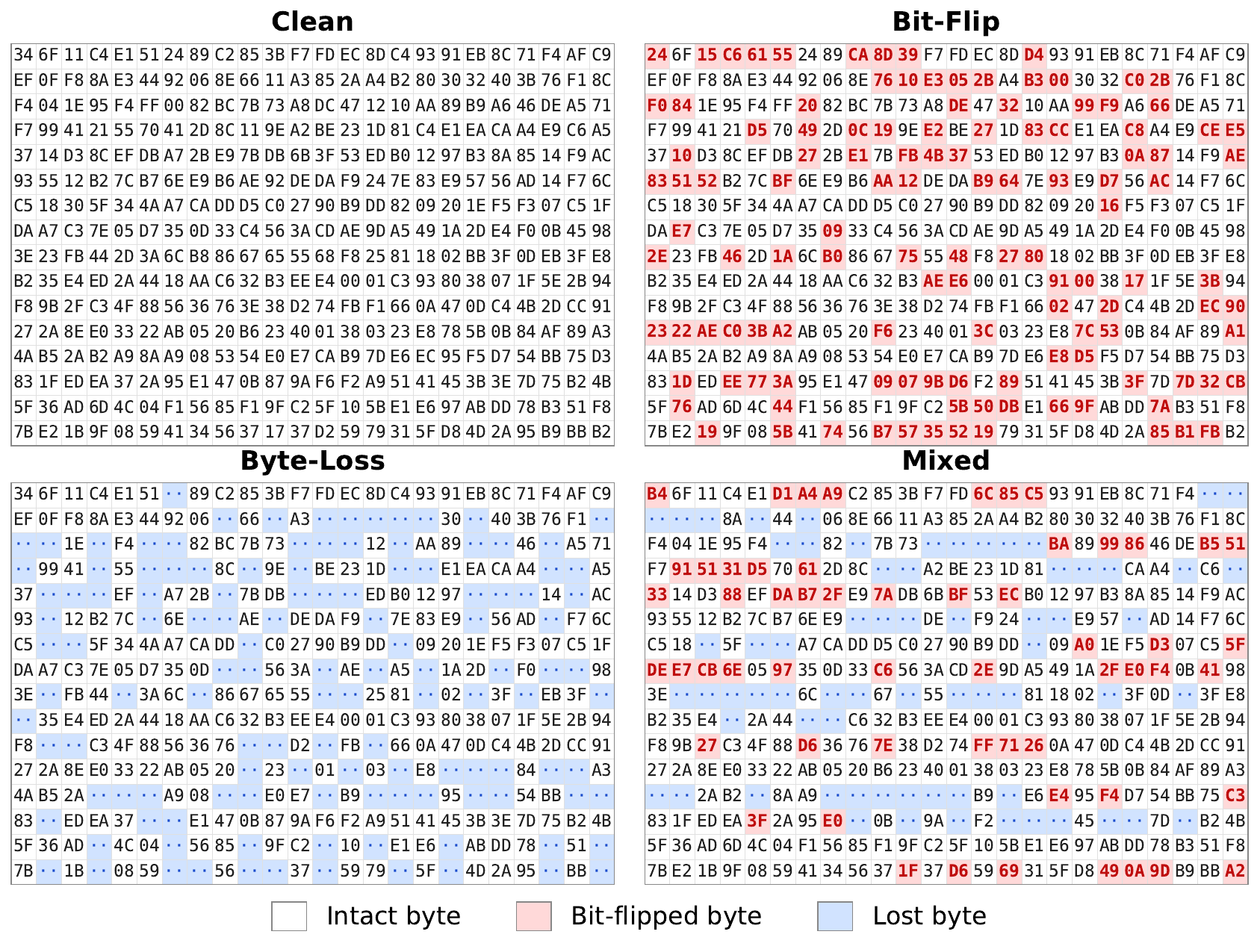}
    \caption{}
    \label{fig:RBCS_b}
  \end{subfigure}

  \caption{The illustration of our proposed Real-world Bitstream Corruption Simulator (RBCS).
(a) The corruption pipeline of bitstream. Bitstream is partitioned into segments and each segment is independently corrupted by Bit-Flip or Byte-Loss.
(b) RBCS applied to a real JPEG bitstream block. Red marks a bit-flipped byte and blue marks a lost byte. }
  \label{fig:RBCS}
\end{figure*}

\section{Problem Formulation and Dataset Construction}\label{sec:benchmark}

\subsection{Problem Formulation}
\label{subsec:problem}

\textbf{Bitstream Action Recognition (BAR)} aims to recognize human actions
directly from compressed image bitstreams without converting them into pixel
images. Let $\mathbf{B} = (b_1, \ldots, b_N)$ denote the byte sequence of a
compressed image bitstream, where each $b_i \in \{0,\ldots,255\}$ is one byte and
$N$ is the stream length.

In practical storage and transmission scenarios, an image bitstream may remain
intact or be corrupted by byte-level disturbances. We denote the actually observed bitstream by
$\mathbf{X}$. If the bitstream is intact, then $\mathbf{X}=\mathbf{B}$. If it is
corrupted, then
\begin{equation}
    \mathbf{X} = \mathcal{R}_{\boldsymbol{\phi}}(\mathbf{B}),
\end{equation}
where $\mathcal{R}_{\boldsymbol{\phi}}$ denotes the corruption process controlled
by the corruption parameter $\boldsymbol{\phi}$, including the corruption type and
severity.

Given the observed image bitstream $\mathbf{X}$, a byte-level recognition model
$\mathcal{F}_\theta$ directly predicts the action label
\begin{equation}
    \hat{y} = \mathcal{F}_\theta(\mathbf{X}), \qquad
    \hat{y} \in \{1, \ldots, C\},
\end{equation}
where $C$ is the number of action classes. During recognition, the model does not
assume whether $\mathbf{X}$ is intact or corrupted.

Unlike conventional action recognition methods that operate on decoded pixel
images, BAR takes the image bitstream itself as input. The model is therefore
required to infer action semantics directly from the byte sequence while remaining
robust to possible bitstream corruption.

\subsection{Real-world Bitstream Corruption Simulator}
Existing corruption benchmarks inject noise in the pixel domain or lower the JPEG
quality factor. Neither reflects how a real channel fails. Transmission errors
strike the compressed byte stream rather than the reconstructed image. A byte can
be flipped or dropped at any position in the file. We therefore model corruption
where it actually occurs. RBCS
operates directly on the encoded byte stream.

Fig.~\ref{fig:RBCS}\subref{fig:RBCS_a} shows the process. RBCS splits the byte
stream into $K$ segments of size $S$. It selects each segment for corruption with
probability $P$. A selected segment receives one of two corruption types. With
probability $p_f$ it undergoes Bit-Flip and with probability $1-p_f$ it undergoes
Byte-Loss. Each byte inside the segment is then corrupted independently with
probability $q$. Under Bit-Flip a corrupted byte has one random bit inverted.
Under Byte-Loss a corrupted byte is deleted from the stream. The four parameters $S$, $P$, $p_f$, and $q$ jointly control the corruption strength
and the balance between the two modes.

Fig.~\ref{fig:RBCS}\subref{fig:RBCS_b} makes the process concrete on a real JPEG
bitstream block. Each cell is one byte in hexadecimal. The panels show the clean
block, Bit-Flip alone, Byte-Loss alone, and the mixed setting that combines both.
Red marks a flipped byte and blue marks a lost byte. The corrupted bytes scatter
across the whole block rather than cluster in one region. The visualization shows
that RBCS perturbs bytes at random positions throughout the stream. This matches
how real corruption can reach any byte of a transmitted file.

\begin{figure*}[!htbp]
    \begin{minipage}[c]{0.30\textwidth}
        \centering
        \begin{subfigure}{\linewidth}
            \includegraphics[width=\linewidth]{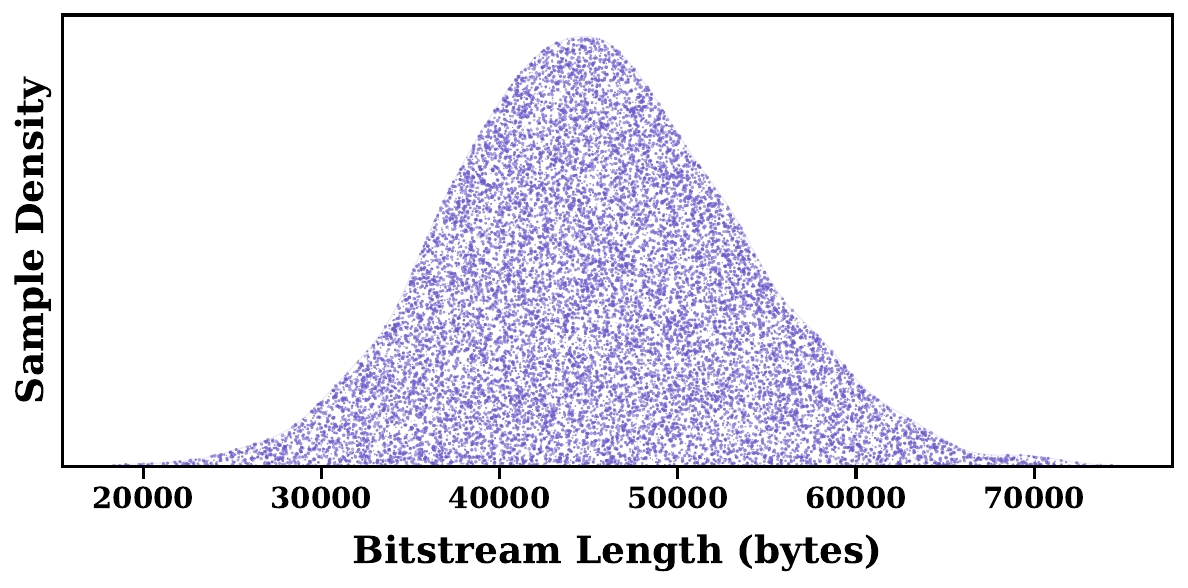}
            \caption{The overall distribution of bitstream lengths.}
        \end{subfigure}
        
        \vspace{0.4cm}
        
        \begin{subfigure}{\linewidth}
            \includegraphics[width=\linewidth]{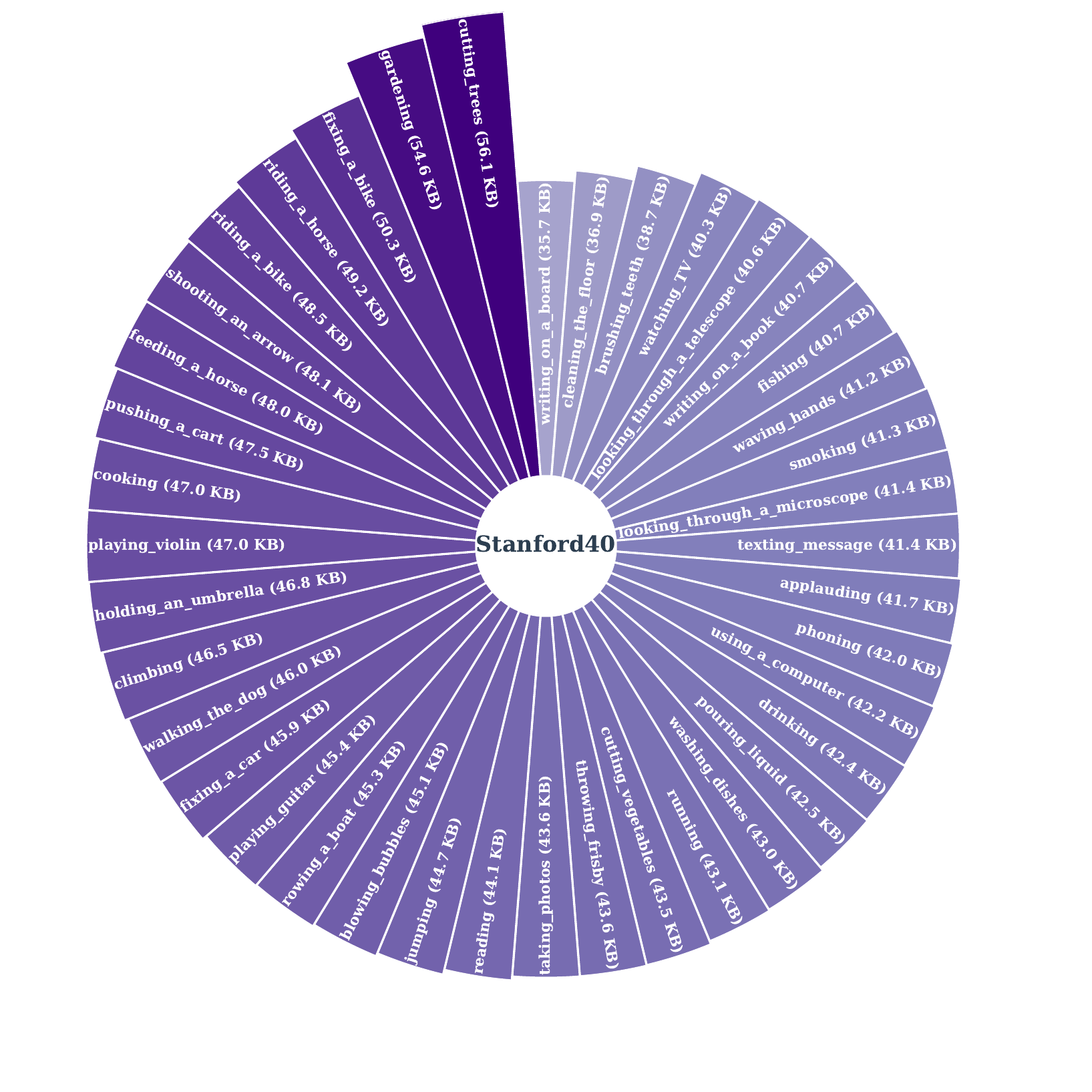}
            \caption{The per-class bitstream lengths on average.}
        \end{subfigure}
        
        \caption{Statistic analysis of bitstream lengths on BAR-Stanford40.}
        \label{fig:bitstream_analysis_stanford}
    \end{minipage}%
    \hfill
    \begin{minipage}[c]{0.69\textwidth}
        \makeatletter
        \def\@captype{table}
        \makeatother
        
        \centering
        \caption{Configuration of the BAR-D corruption benchmark and the resulting decode rates. Each scenario is fixed by an RBCS tuple $(S, P, p_f, q)$, and $\rho \approx P\cdot q$ is the expected fraction of corrupted bytes. The decode rate is the share of corrupted streams that a standard decoder can still recover into an image.}
        \label{tab:benchmark}
        \renewcommand{\arraystretch}{1.5}
        \setlength{\tabcolsep}{6pt} 
        
        \small
        \begin{tabular}{l c c c c c c c}
        \toprule
        \multirow{2}{*}{\textbf{Scenario}} & \multirow{2}{*}{$S$} & \multirow{2}{*}{$P$} & \multirow{2}{*}{$p_f$} & \multirow{2}{*}{$q$} & \multirow{2}{*}{$\rho$ (\%)} & \multicolumn{2}{c}{\textbf{Decode Rate (\%)}} \\
        \cmidrule(lr){7-8}
         & & & & & & BAR-Stanford40 & BAR-PPMI \\
        \midrule
        Clean        & --  & --   & --  & --   & 0     & 100.0 & 100.0 \\
        \midrule
        Light-Flip   & 128 & 0.35 & 1.0 & 0.35 & 12.25 & 0.45  & 0.62  \\
        Light-Loss   & 128 & 0.35 & 0.0 & 0.35 & 12.25 & 1.07  & 1.14  \\
        Light-Mixed  & 128 & 0.35 & 0.5 & 0.35 & 12.25 & 0.58  & 0.62  \\
        \midrule
        Medium-Flip  & 64  & 0.55 & 1.0 & 0.55 & 30.25 & 0.04  & 0.05  \\
        Medium-Loss  & 64  & 0.55 & 0.0 & 0.55 & 30.25 & 0.05  & 0.03  \\
        Medium-Mixed & 64  & 0.55 & 0.5 & 0.55 & 30.25 & 0.05  & 0.10  \\
        \midrule
        Heavy-Flip   & 32  & 0.65 & 1.0 & 0.65 & 42.25 & 0.00  & 0.00  \\
        Heavy-Loss   & 32  & 0.65 & 0.0 & 0.65 & 42.25 & 0.00  & 0.00  \\
        Heavy-Mixed  & 32  & 0.65 & 0.5 & 0.65 & 42.25 & 0.00  & 0.00  \\
        \midrule
        Extreme-Flip & 16  & 0.75 & 1.0 & 0.75 & 56.25 & 0.00  & 0.00  \\
        Extreme-Loss & 16  & 0.75 & 0.0 & 0.75 & 56.25 & 0.00  & 0.00  \\
        Extreme-Mixed& 16  & 0.75 & 0.5 & 0.75 & 56.25 & 0.00  & 0.00  \\
        \bottomrule
        \end{tabular}
    \end{minipage}
\end{figure*}

\subsection{Construction of BAR-D}
\label{subsec:dataset}

Building on RBCS, we construct BAR-D, the first large-scale dataset for bitstream
action recognition under corruption. We build it on two widely used action
recognition sources, Stanford40 and PPMI, which yield the subsets BAR-Stanford40
and BAR-PPMI. To obtain a uniform input we canonicalize every image through one
encoding pipeline. We resize each image so its shorter side is 256 pixels,
center-crop it to $224\times224$, and encode it as a JPEG bitstream at quality 100.
We then cap each bitstream to $L_{\max}=50{,}000$ bytes, padding shorter streams and
truncating longer ones so that all samples share a common length.

Before padding or truncation, we analyze the original canonical JPEG lengths to verify whether file size provides a trivial shortcut. Fig.~\ref{fig:bitstream_analysis_stanford} analyzes the bitstream lengths on
BAR-Stanford40, with the corresponding BAR-PPMI analysis in Fig.~9 of the
supplementary material. The top row shows that the lengths concentrate in a narrow
band around a median near 44~KB. The bottom row shows that the per-class average
lengths fall within a small range. Bitstream length therefore carries little class
information and offers no shortcut for recognition.

We then apply RBCS to the canonicalized bitstreams to build a graded suite of
evaluation scenarios, shown in Table~\ref{tab:benchmark}. The suite spans four
severity levels named Light, Medium, Heavy, and Extreme. From Light to Extreme the
segment size $S$ shrinks from 128 to 16 while $P$ and $q$ rise together, so the
expected corruption ratio $\rho$ grows from about 12\% to 56\%. Each level appears
in three corruption types set by $p_f$, namely Bit-Flip ($p_f=1$), Byte-Loss
($p_f=0$), and Mixed ($p_f=0.5$). With an uncorrupted Clean scenario, BAR-D defines
13 evaluation scenarios.

The last two columns of Table~\ref{tab:benchmark} report the decode rates at which a standard
decoder recovers an image from the corrupted stream. Because RBCS scatters
corruption across the whole bitstream, a damaged byte almost always lands early in the
entropy-coded data and desynchronizes everything after it. The decode rate
therefore collapses to around 1\% already at the Light level and reaches nearly
zero from Medium onward.

\begin{figure*}[t]  
  \centering
  \includegraphics[width=1.0\textwidth]{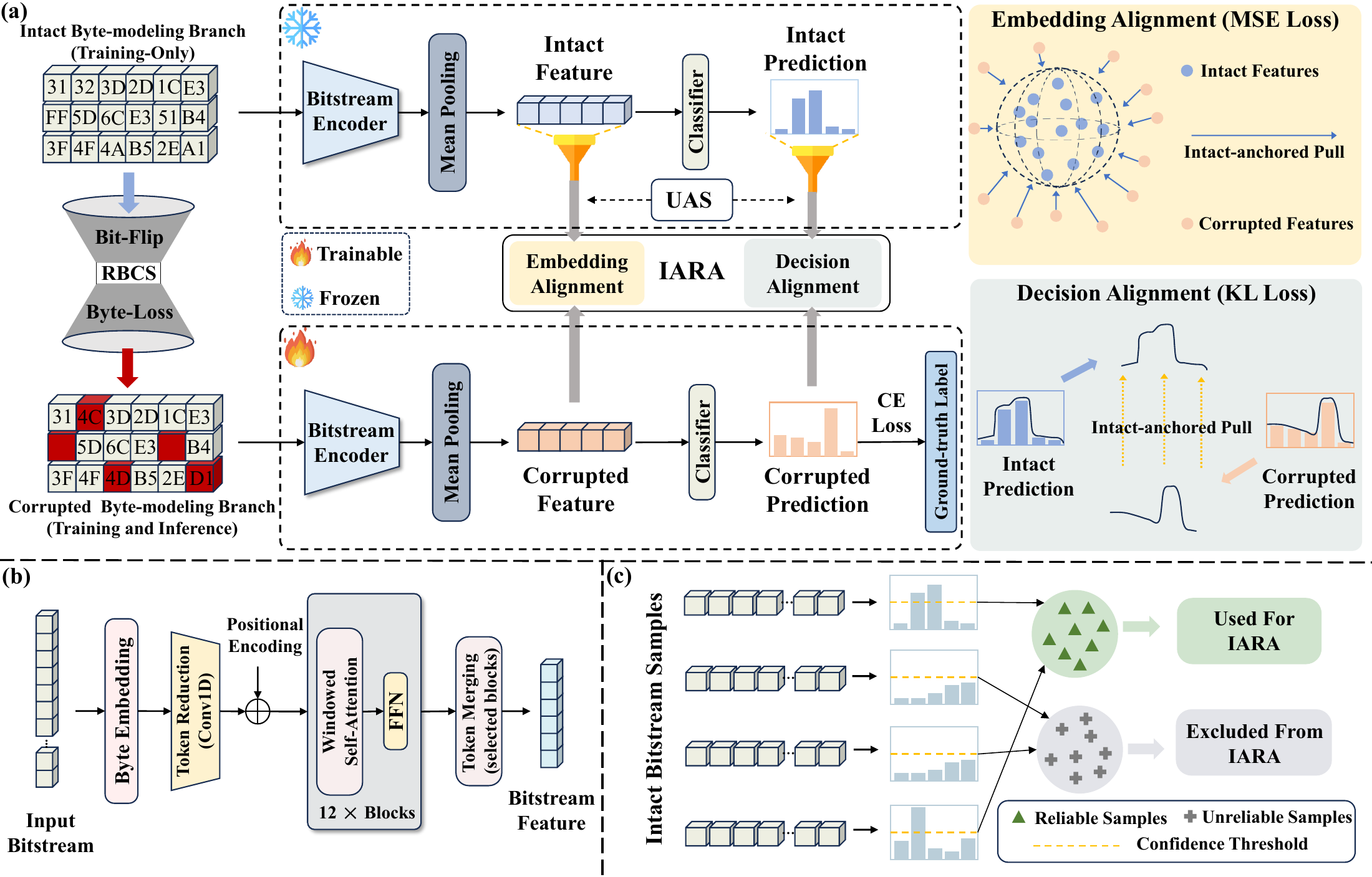} 
  \caption{Overview of our proposed BRACE framework.
(a) BRACE pairs two byte-modeling branches. The intact
branch is frozen and encodes the clean bitstream into anchor representations. The corrupted branch is trainable and encodes the corrupted bitstream. Intact-Anchored Representation Alignment (IARA) pulls the corrupted feature and prediction toward these
anchors through embedding alignment (MSE loss) and decision alignment (KL loss). A cross-entropy loss on the ground-truth label trains the corrupted branch. 
(b) The bitstream encoder adopts the ByteFormer architecture~\cite{hortonbytes}. It embeds the raw bytes and reduces the token count with a convolution. Positional encoding is then added before a stack of windowed self-attention blocks. These blocks merge tokens at selected depths to shorten the sequence. Mean pooling yields the final representation.
(c) Unreliable-Anchor Suppression (UAS) aims to suppress unreliable intact samples whose confidence is below a threshold. IARA aligns only to reliable anchors.}
  \label{fig:framework}
\end{figure*}

\section{The Proposed BRACE Framework}
\label{sec:method}

\subsection{Overview}
\label{subsec:overview}

BRACE recognizes an action directly from a corrupted bitstream by aligning its
representation with that of the corresponding clean bitstream. As shown in
Fig.~\ref{fig:framework}(a), the framework consists of two byte-modeling branches that
share the same encoder architecture. The \emph{intact branch} encodes the clean
bitstream $\mathbf{B}$ and is kept frozen, providing stable anchor representations. The \emph{corrupted branch} encodes the corrupted bitstream
$\widetilde{\mathbf{B}}=\mathcal{R}_{\boldsymbol{\phi}}(\mathbf{B})$ produced by RBCS
during training. This branch is deployed at inference time and takes the observed image
bitstream $\mathbf{X}$ as input, where $\mathbf{X}$ may be intact or corrupted. Both
branches are initialized from the same encoder pretrained on clean bitstreams.

Each branch maps its input to a pooled feature and a class distribution. For the
intact branch,
\begin{equation}
\mathbf{z}^{\mathrm{I}}=f^{\mathrm{I}}(\mathbf{B}),\qquad
\mathbf{p}^{\mathrm{I}}=\mathrm{softmax}\!\big(g^{\mathrm{I}}(\mathbf{z}^{\mathrm{I}})\big),
\end{equation}
and $(\mathbf{z}^{\mathrm{C}},\mathbf{p}^{\mathrm{C}})$ are defined analogously for
the corrupted branch from $\widetilde{\mathbf{B}}$, where $f$ is the bitstream
encoder (Section~\ref{subsec:encoder}) and $g$ a linear classifier. Corruption
shifts $\mathbf{z}^{\mathrm{C}}$ and $\mathbf{p}^{\mathrm{C}}$ away from their
clean counterparts. BRACE corrects this shift through Intact-Anchored
Representation Alignment, which pulls the corrupted feature and prediction
toward the intact anchors, applied only where the anchor is reliable. Only the
corrupted branch is updated. The intact branch supplies fixed targets.

\subsection{Bitstream Encoder}
\label{subsec:encoder}

Both branches adopt a ByteFormer\cite{hortonbytes} encoder, shown in
Fig.~\ref{fig:framework}(b). The input bitstream is first mapped to byte
embeddings, whose length is reduced by a one-dimensional convolution to form a
compact token sequence. After adding positional encodings, the tokens pass
through a stack of transformer blocks, each combining windowed self-attention
with a feed-forward network, and token merging is applied at selected blocks to
progressively shorten the sequence. The resulting tokens are averaged by mean
pooling into a single feature vector, which the classifier maps to class logits.

\subsection{Intact-Anchored Representation Alignment}
\label{subsec:iara}

IARA aligns the corrupted branch with the intact anchors at two complementary levels,
embedding and decision, and restricts alignment to reliable anchors through
Unreliable-Anchor Suppression.

\noindent\textbf{Embedding Alignment.}
To restore the clean geometry of the corrupted feature, we minimize the squared
distance between the two pooled features,
\begin{equation}
\mathcal{L}_{\mathrm{feat}}=\frac{1}{|\mathcal{M}|}\sum_{i\in\mathcal{M}}
\big\|\,\mathbf{z}^{\mathrm{C}}_i-\mathrm{sg}(\mathbf{z}^{\mathrm{I}}_i)\,\big\|_2^2,
\end{equation}
where $\mathrm{sg}(\cdot)$ is the stop-gradient operator and $\mathcal{M}$ is the
set of reliable samples defined below. The intact feature acts as a fixed anchor,
so gradients flow only into the corrupted branch.

\noindent\textbf{Decision Alignment.}
We further align the predicted distributions so that the two branches reach
consistent decisions. With temperature-scaled distributions
$\mathbf{p}^{\tau}=\mathrm{softmax}(\cdot/\tau)$, we minimize
\begin{equation}
\mathcal{L}_{\mathrm{pred}}=\frac{\tau^2}{|\mathcal{M}|}\sum_{i\in\mathcal{M}}
\mathrm{KL}\!\big(\mathrm{sg}(\mathbf{p}^{\mathrm{I},\tau}_i)\,\big\|\,
\mathbf{p}^{\mathrm{C},\tau}_i\big),
\end{equation}
where $\tau$ softens both distributions so that the anchor's full class structure,
beyond its top class, informs the alignment, and the $\tau^2$ factor keeps the
gradient magnitude stable across temperatures.

\noindent\textbf{Unreliable-Anchor Suppression.}
The anchor is reliable only when the intact branch is confident. Aligning to an uncertain
anchor can mislead the corrupted branch. We therefore suppress anchors whose top
probability falls below a threshold $\gamma$ and keep only the reliable ones,
\begin{equation}
\mathcal{M}=\big\{\, i:\ \max_{c}\,\mathbf{p}^{\mathrm{I}}_{i,c}>\gamma \,\big\}.
\end{equation}
As illustrated in Fig.~\ref{fig:framework}(c), samples with a peaked anchor distribution
are deemed reliable and used for IARA, while those with a flat distribution are
suppressed. Both alignment terms use the same set $\mathcal{M}$.

\subsection{Training Objective}
\label{subsec:objective}

BRACE is trained by minimizing a single objective that combines a classification
loss with the two alignment terms,
\begin{equation}
\mathcal{L}=\mathcal{L}_{\mathrm{ce}}
+\lambda_{\mathrm{feat}}\,\mathcal{L}_{\mathrm{feat}}
+\lambda_{\mathrm{pred}}\,\mathcal{L}_{\mathrm{pred}},
\end{equation}
where the classification loss
\begin{equation}
\mathcal{L}_{\mathrm{ce}}=\mathrm{CE}\!\big(\mathbf{p}^{\mathrm{C}},\,y\big)
\end{equation}
is the cross-entropy between the corrupted prediction and the ground-truth label
$y$. The three terms play complementary roles. $\mathcal{L}_{\mathrm{ce}}$ makes
the corrupted branch discriminative for the task, $\mathcal{L}_{\mathrm{feat}}$
restores the clean geometry of its feature, and $\mathcal{L}_{\mathrm{pred}}$
aligns its decision with the anchor. Gradients update only the corrupted branch
and its classifier, while the intact branch stays frozen. At test time, only the
corrupted branch is evaluated. The intact branch and the alignment terms are used
solely during training.

\section{Experiments}\label{sec:experiments}

\begin{table*}[t]
\centering
\caption{Model Comparisons on BAR-Stanford40 and BAR-PPMI subsets. Accuracy (\%) and mAP (\%) are used for measuring the performance on Clean data and corrupted data at 4 severity levels. These metric values are computed by averaging the Flip, Loss, Mixed corruption.
``Corrupt Avg'' is the mean over all scenarios. 
The best results are in \textbf{bold}.}
\label{tab:main_results}
\resizebox{\textwidth}{!}{%
\setlength{\tabcolsep}{4.5pt}
\renewcommand{\arraystretch}{1.3}
\begin{tabular}{l|c|cc|cc|cc|cc|cc|cc}
\hline
\multicolumn{1}{c|}{\multirow{2}{*}{Method}} & \multicolumn{1}{c|}{\multirow{2}{*}{Domain}} & \multicolumn{2}{c|}{Clean} & \multicolumn{2}{c|}{Light Avg} & \multicolumn{2}{c|}{Medium Avg} & \multicolumn{2}{c|}{Heavy Avg} & \multicolumn{2}{c|}{Extreme Avg} & \multicolumn{2}{c}{Corrupt Avg} \\
\cline{3-14}
 & & Acc & mAP & Acc & mAP & Acc & mAP & Acc & mAP & Acc & mAP & Acc & mAP \\
\hline
\multicolumn{14}{c}{\textbf{BAR-Stanford40}} \\
\hline
ResNet-50                      & Pixel      & 85.30 & 90.09 & 0.02  & 2.57  & 0.00  & 2.52  & 0.00  & 2.50  & 0.00  & 2.50  & 0.01  & 2.52 \\
ViT-B/16                       & Pixel      & 83.33 & 86.23 & 0.00  & 2.59  & 0.00  & 2.51  & 0.00  & 2.50  & 0.00  & 2.50  & 0.00  & 2.52 \\
Swin-Tiny                      & Pixel      & 85.52 & 90.08 & 0.01  & 2.58  & 0.00  & 2.52  & 0.00  & 2.50  & 0.00  & 2.50  & 0.00  & 2.52 \\
ConvNeXt-Tiny                  & Pixel      & \textbf{87.55} & \textbf{92.66} & 0.02  & 2.58  & 0.00  & 2.52  & 0.00  & 2.50  & 0.00  & 2.50  & 0.00  & 2.52 \\
DeiT-Small                     & Pixel      & 81.36 & 85.94 & 0.00  & 2.57  & 0.00  & 2.52  & 0.00  & 2.50  & 0.00  & 2.50  & 0.00  & 2.52 \\
\hline
Optimal Codebook               & Compressed & 12.51 & 10.30 & 0.09  & 2.74  & 0.01  & 2.72  & 0.00  & 2.50  & 0.00  & 2.50  & 0.02  & 2.61 \\
Direct Feature Extractor       & Compressed & 19.14 & 15.28 & 0.13  & 2.55  & 0.01  & 2.51  & 0.00  & 2.50  & 0.00  & 2.50  & 0.04  & 2.52 \\
Transform-Bitstream Classifier & Compressed & 29.32 & 28.34 & 0.20  & 2.85  & 0.01  & 2.62  & 0.00  & 2.50  & 0.00  & 2.50  & 0.05  & 2.62 \\
\hline
bGPT                           & Bitstream  & 6.54  & 4.52  & 4.86  & 3.97  & 4.19  & 3.80  & 4.04  & 3.67  & 3.83  & 3.54  & 4.23  & 3.74 \\
MEGABYTE                       & Bitstream  & 4.75  & 4.08  & 4.19  & 3.84  & 3.93  & 3.82  & 3.75  & 3.85  & 3.55  & 3.74  & 3.85  & 3.81 \\
MambaByte                      & Bitstream  & 3.38  & 3.03  & 3.38  & 3.00  & 3.38  & 3.07  & 3.38  & 3.05  & 3.38  & 3.03  & 3.38  & 3.04 \\
MBLM                           & Bitstream  & 3.24  & 3.54  & 3.07  & 3.51  & 2.77  & 3.51  & 2.66  & 3.50  & 2.62  & 3.50  & 2.78  & 3.51 \\
ByteFormer                     & Bitstream  & 61.33 & 63.84 & 53.38 & 55.12 & 25.64 & 25.34 & 12.50 & 12.16 & 5.91  & 5.53  & 24.36 & 24.54 \\
\textbf{BRACE (Ours)  }            & Bitstream  & 62.95 & 66.55 & \textbf{59.15} & \textbf{61.56} & \textbf{45.70} & \textbf{45.96} & \textbf{32.01} & \textbf{31.30} & \textbf{18.15} & \textbf{16.78} & \textbf{38.75} & \textbf{38.90} \\
\hline
\multicolumn{14}{c}{\textbf{BAR-PPMI}} \\
\hline
ResNet-50                      & Pixel      & 71.65 & 77.21 & 0.07  & 4.30  & 0.00  & 4.22  & 0.00  & 4.17  & 0.00  & 4.17  & 0.02  & 4.22 \\
ViT-B/16                       & Pixel      & 64.84 & 66.63 & 0.00  & 4.31  & 0.00  & 4.21  & 0.00  & 4.17  & 0.00  & 4.17  & 0.00  & 4.22 \\
Swin-Tiny                      & Pixel      & 67.03 & 73.02 & 0.00  & 4.35  & 0.00  & 4.21  & 0.00  & 4.17  & 0.00  & 4.17  & 0.00  & 4.22 \\
ConvNeXt-Tiny                  & Pixel      & \textbf{74.56} & \textbf{81.24} & 0.07  & 4.35  & 0.00  & 4.21  & 0.00  & 4.17  & 0.00  & 4.17  & 0.02  & 4.23 \\
DeiT-Small                     & Pixel      & 65.75 & 70.95 & 0.05  & 4.34  & 0.00  & 4.21  & 0.00  & 4.17  & 0.00  & 4.17  & 0.01  & 4.22 \\
\hline
Optimal Codebook               & Compressed & 9.29  & 7.61  & 0.07  & 4.28  & 0.00  & 4.19  & 0.00  & 4.17  & 0.00  & 4.17  & 0.02  & 4.20 \\
Direct Feature Extractor       & Compressed & 7.81  & 6.95  & 0.06  & 4.25  & 0.00  & 4.19  & 0.00  & 4.17  & 0.00  & 4.17  & 0.02  & 4.20 \\
Transform-Bitstream Classifier & Compressed & 17.53 & 16.69 & 0.13  & 4.45  & 0.01  & 4.22  & 0.00  & 4.17  & 0.00  & 4.17  & 0.04  & 4.25 \\
\hline
bGPT                           & Bitstream  & 12.15 & 11.15 & 9.86  & 10.06 & 9.00  & 9.50  & 8.00  & 9.05  & 7.43  & 8.68  & 8.58  & 9.32 \\
MEGABYTE                       & Bitstream  & 12.29 & 11.51 & 10.56 & 10.61 & 9.27  & 9.50  & 8.24  & 9.07  & 7.32  & 8.42  & 8.85  & 9.40 \\
MambaByte                      & Bitstream  & 6.38  & 6.18  & 6.32  & 6.24  & 5.75  & 6.31  & 5.43  & 6.39  & 4.70  & 6.37  & 5.55  & 6.33 \\
MBLM                           & Bitstream  & 12.91 & 11.25 & 12.34 & 11.20 & 12.10 & 11.14 & 11.94 & 11.10 & 10.45 & 10.52 & 11.71 & 10.99 \\
ByteFormer                     & Bitstream  & 50.07 & 52.37 & 37.19 & 40.20 & 15.28 & 16.42 & 8.23  & 9.03  & 4.99  & 6.14  & 16.42 & 17.95 \\
\textbf{BRACE (Ours)   }           & Bitstream  & 51.02 & 53.01 & \textbf{45.37} & \textbf{47.53} & \textbf{30.01} & \textbf{31.34} & \textbf{18.90} & \textbf{19.06} & \textbf{12.15} & \textbf{12.13} & \textbf{26.61} & \textbf{27.52} \\
\hline
\end{tabular}%
}
\end{table*}

\subsection{Experimental Settings}

\subsubsection{Evaluation Metrics and Protocol}
We evaluate every method on BAR-D (Section~\ref{subsec:dataset}) across its 13
scenarios. We report Top-1 Accuracy (\%) and mean Average Precision (mAP, \%) for
each scenario. We summarize robustness with the Corrupt Average. It is the mean
over the 12 corrupted scenarios. For decoding-based methods, a sample that fails to
decode counts as a misclassification with zero confidence. All corruption tests use
a fixed random seed for reproducibility.

\subsubsection{State-of-the-Art Comparison Methods}
We compare methods from the pixel, compressed, and bitstream domains.

\paragraph{Pixel-domain methods} These methods follow the standard full-decode paradigm and
reconstruct RGB pixels before feature extraction. We evaluate
ResNet-50~\cite{he2016deep}, ViT-B/16~\cite{dosovitskiy2021an}, Swin-Tiny~\cite{liu2021swin},
ConvNeXt-Tiny~\cite{liu2022convnet}, and DeiT-Small~\cite{touvron2021training}.

\paragraph{Compressed-domain methods} These methods operate on partially decoded
representations such as DCT coefficients. They avoid full
spatial reconstruction but still require structural parsing of the bitstream. We
evaluate Optimal Codebook~\cite{jamil2019optimal}, Direct Feature
Extractor~\cite{shen1996direct}, and Transform-Bitstream Classifier
(TBC)~\cite{hill2021transform}.

\paragraph{Bitstream-domain methods} These methods operate directly on raw bytes without
decoding or parsing. We compare
bGPT~\cite{wu2024beyond}, MEGABYTE~\cite{yu2023MEGABYTE}, MambaByte~\cite{wang2024MambaByte},
MBLM~\cite{egli2025multiscale}, and ByteFormer~\cite{hortonbytes}.

\subsubsection{Implementation Details}
All methods follow one protocol. We apply RBCS once to each canonical bitstream
with a fixed seed. Each domain then consumes the corrupted stream in its native
form, as RGB pixels, DCT coefficients, or raw bytes truncated or zero-padded to
$L_{\max}=50{,}000$. Pixel-domain models start from ImageNet weights and are
fine-tuned per dataset with AdamW (learning rate $1\times10^{-4}$, weight decay
$0.05$). Compressed-domain models use the configurations from their original papers.
The bitstream baselines bGPT, MEGABYTE, MambaByte, and MBLM follow their official
codebases. ByteFormer and BRACE adopt the Tiny configuration ($d{=}192$, 12 blocks,
$k{=}8$). We fine-tune BRACE end-to-end with AdamW (learning rate $3\times10^{-5}$,
weight decay $0.05$), starting both branches from ByteFormer weights pretrained on
clean bitstreams. We freeze the intact branch and train only the corrupted branch.
During training the corrupted branch sees an RBCS-corrupted bitstream with
probability $p_{\mathrm{aug}}=0.7$ and the clean bitstream otherwise. IARA uses
$\tau=4$, $\gamma=0.5$, and $\lambda_{\mathrm{feat}}=\lambda_{\mathrm{pred}}=1$. All
experiments run on a single NVIDIA RTX 4090 GPU.

\begin{figure*}[!htbp] 
  \centering
  \includegraphics[width=\linewidth]{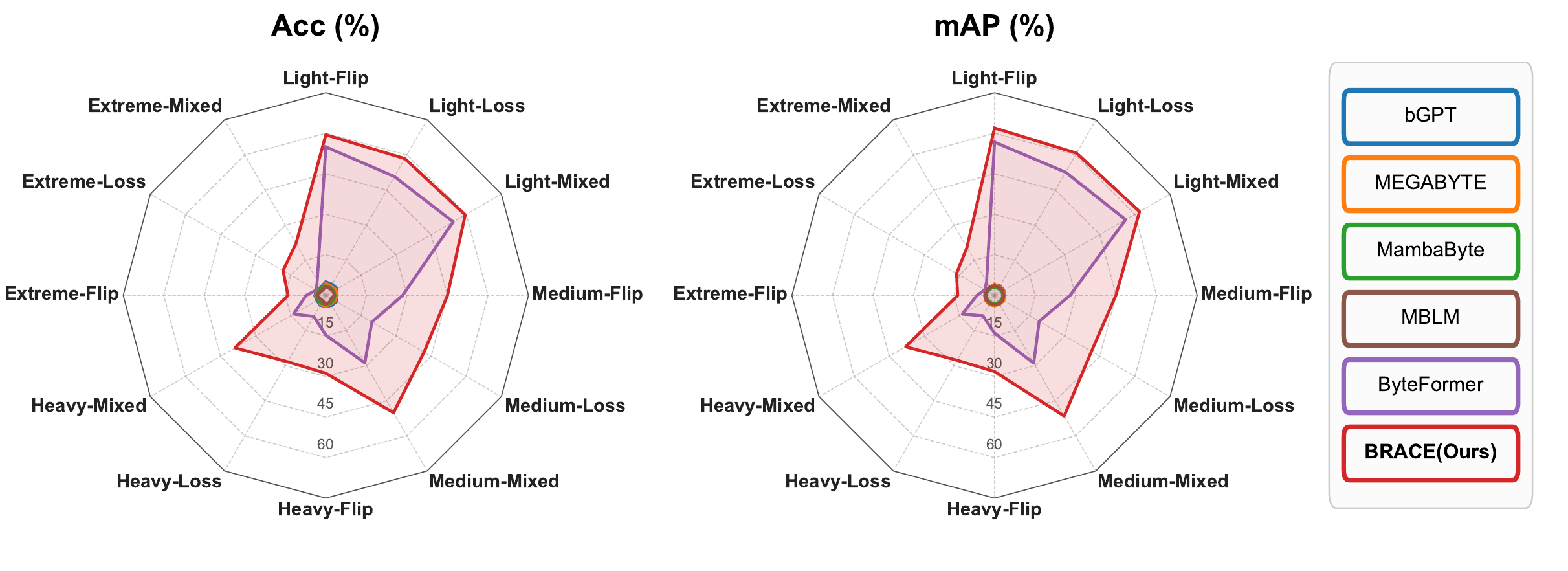}
  
  \caption{Per-scenario comparison of bitstream-domain methods on BAR-Stanford40 in terms of Accuracy (\%) and mAP (\%). Each axis represents one of the 12 corruption scenarios. Our BRACE outperforms all comparison methods across all scenarios.}
  \label{fig:12_scenario_stanford}
\end{figure*}

\begin{figure*}[!htbp] 
  \centering
  \includegraphics[width=\linewidth]{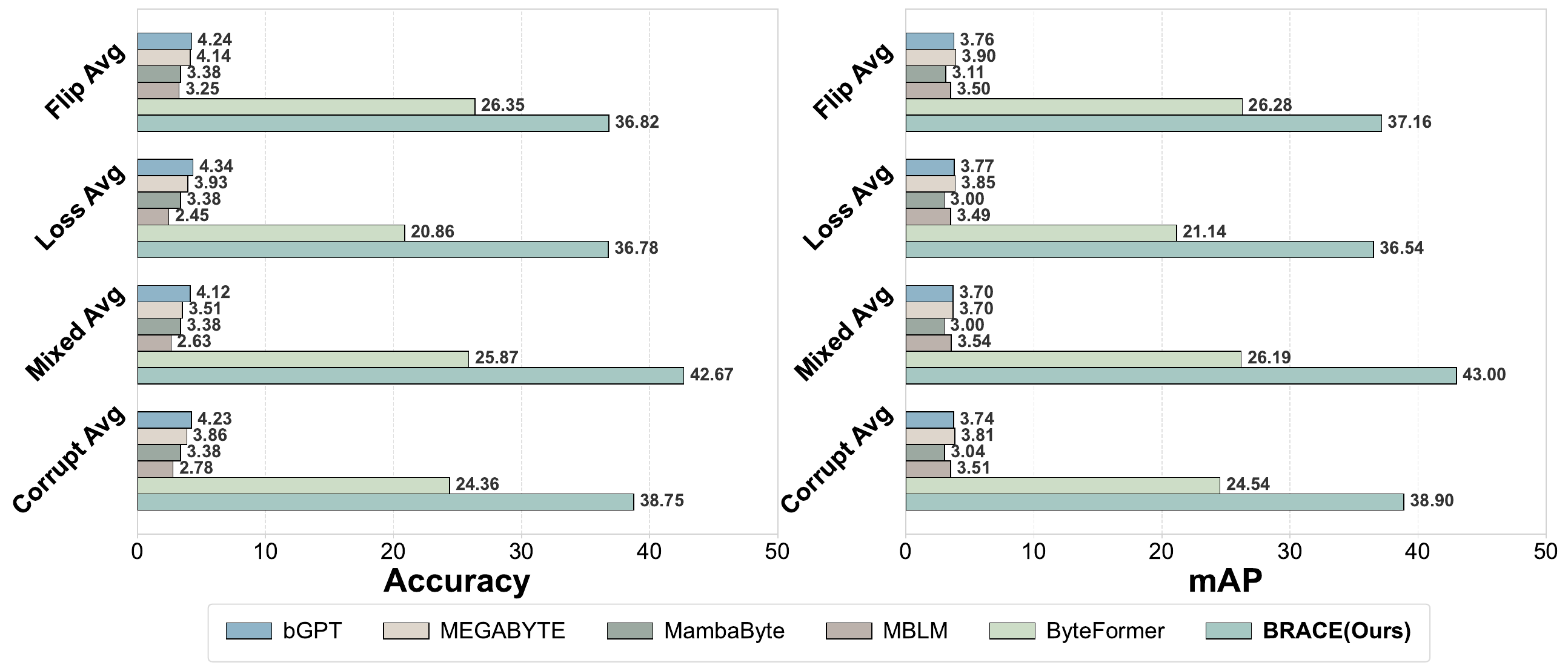}
  
  \caption{Comparison of bitstream-domain methods by corruption type on BAR-Stanford40. The results report the average Accuracy and mAP for four severity levels on different corruption types. Our BRACE outperforms all comparison methods across all corruption types.}
  \label{fig:Different_Corrupt_type_stanford}
\end{figure*}

\begin{figure*}[!htbp] 
  \centering
  \includegraphics[width=\linewidth]{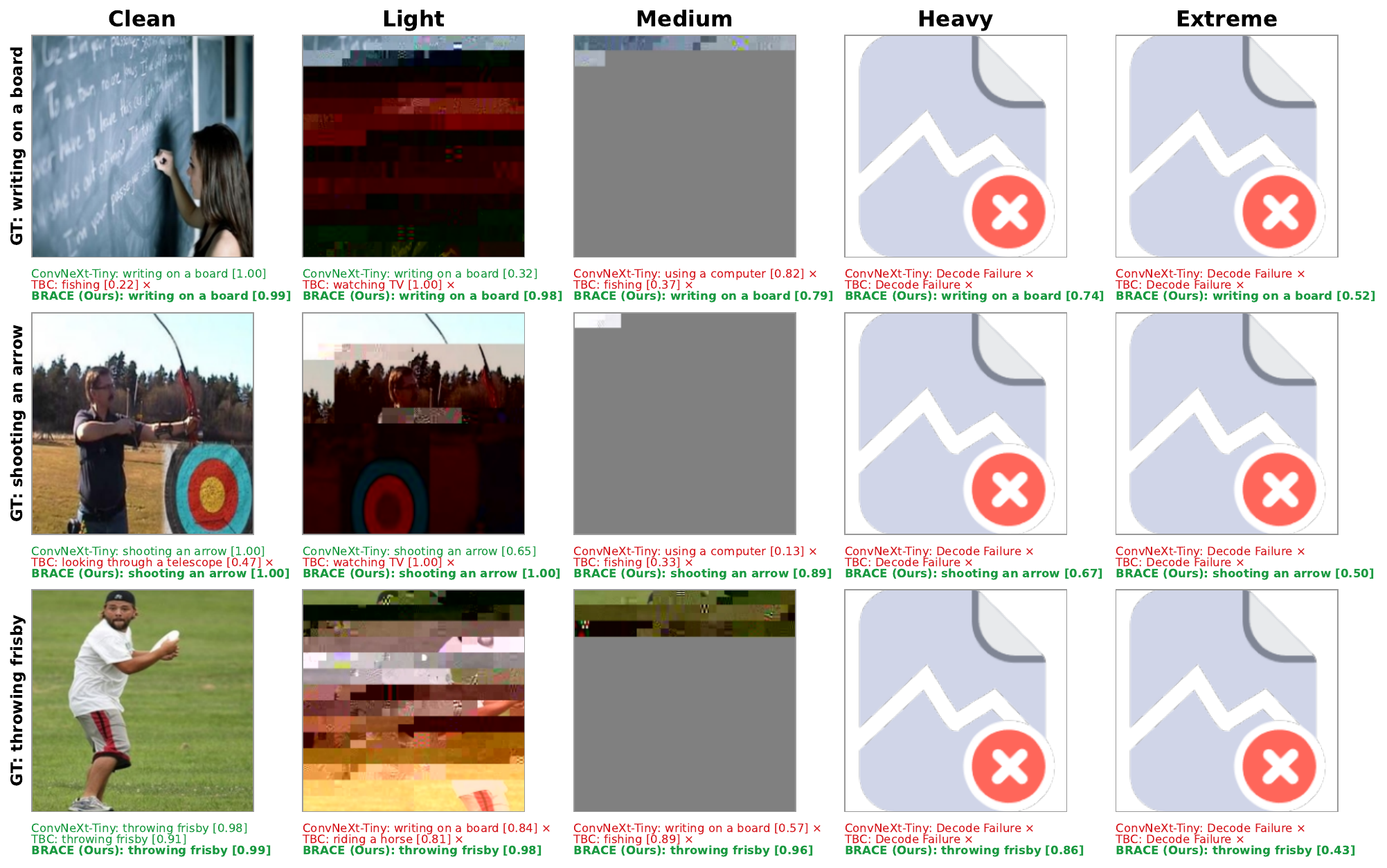}
  
  \caption{Illustration of Qualitative comparison on BAR-Stanford40. Each row is one image and each column represents its severity. The ConvNeXt-Tiny, the Transform-Bitstream Classifier (TBC), and our BRACE from three domains give their predictions. The correct predictions are in green and wrong or failed predictions are in red. Pixel-domain and compressed-domain methods fail once the bitstream decoding fails, while BRACE keeps recognizing the action from the raw bytes.
}
  \label{fig:qualitative}
\end{figure*}

\subsection{Comparison with the State-of-the-Arts}
\label{subsec:comparison}

\subsubsection{Overall Performance}
Table~\ref{tab:main_results} reports the overall results on BAR-Stanford40 and BAR-PPMI.
We make three observations.

\noindent\textbf{Pixel-domain and compressed-domain methods collapse under corruption.}
Pixel-domain models reach the highest clean accuracy but drop to near-random once
corruption begins. Compressed-domain models fail the same way. Both rely on a parsable
JPEG structure that the corrupted stream no longer provides. The collapse holds across
every architecture and severity level. It points to decoding as the bottleneck rather
than model capacity.

\noindent\textbf{Bitstream-domain methods stay functional but vary widely.}
Byte-level models skip decoding and keep non-trivial accuracy at every corruption level.
Their strength still varies sharply. The generalist models bGPT, MEGABYTE, MambaByte,
and MBLM stay weak even on clean inputs. ByteFormer is built for byte-level visual
understanding and forms a much stronger baseline.

\noindent\textbf{BRACE achieves the best robustness on both subsets.}
BRACE reaches the highest Corrupt Average by a clear margin and ranks first at every
corrupted level. On clean inputs it stays slightly above the ByteFormer baseline, so
robustness does not come at the expense of clean accuracy.

\subsubsection{Per-Scenario Comparison among Bitstream-Domain Methods}
Pixel-domain and compressed-domain methods barely function under corruption. We
therefore compare the bitstream-domain methods scenario by scenario.
Fig.~\ref{fig:12_scenario_stanford} plots Accuracy and mAP across all 12
scenarios on BAR-Stanford40. BRACE covers the largest area in both plots. It stays
reliable across all corruption scenarios. The same pattern holds on BAR-PPMI, shown
in Fig.~11 of the supplementary material.

\begin{figure}
  \centering
  \includegraphics[width=1\linewidth]{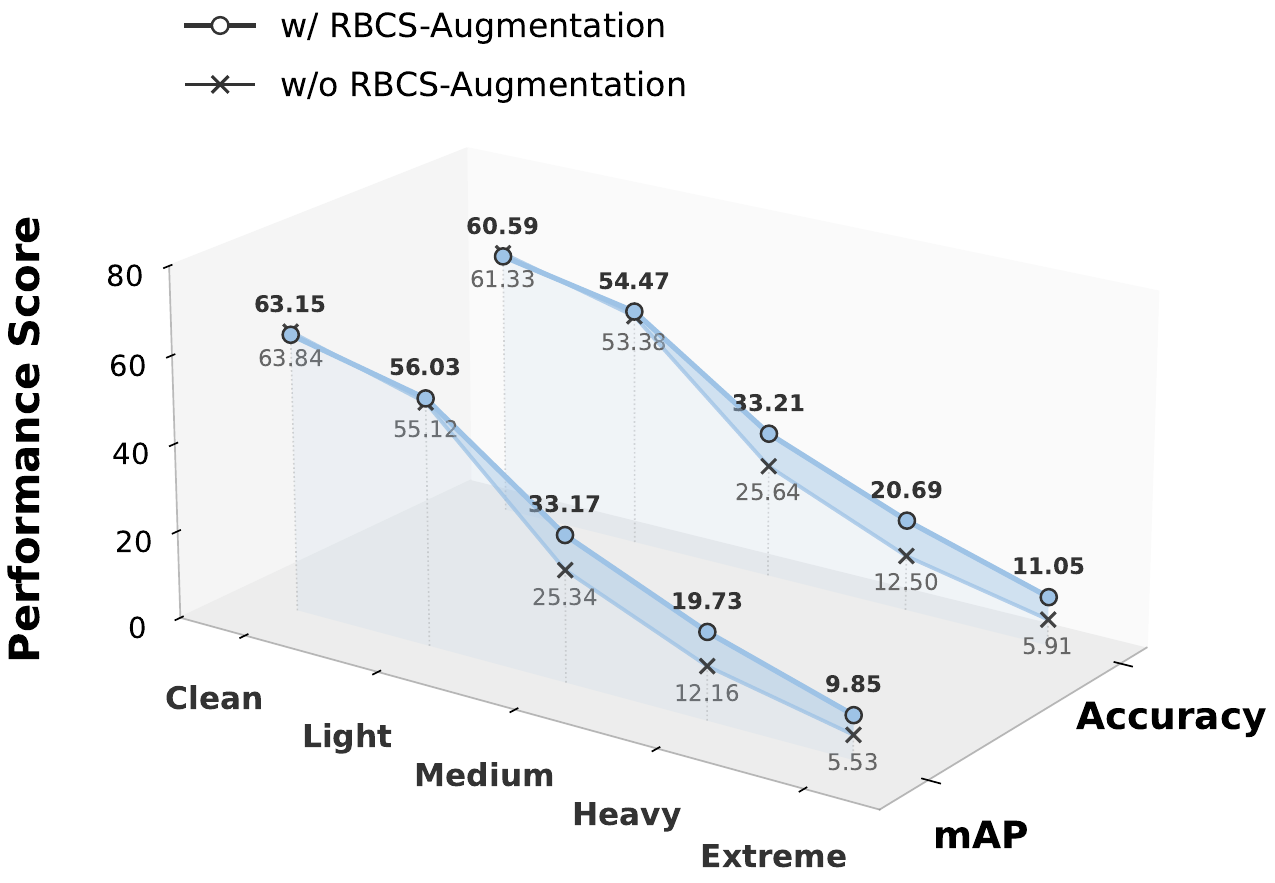}
  \caption{Ablation on the effectiveness of RBCS augmentation on BAR-Stanford40.
  }\label{fig:rbcs_ablation}
\end{figure}

\subsubsection{Analysis by Corruption Type}
Fig.~\ref{fig:Different_Corrupt_type_stanford} breaks the Corrupt Average down by
corruption type on BAR-Stanford40. BRACE gives the best Accuracy and mAP under Flip, Loss, and Mixed. It favors no
particular corruption mechanism. BAR-PPMI shows the same behavior, reported in
Fig.~12 of the supplementary material.
The three types differ in difficulty. At equal severity Mixed yields the highest
accuracy. It splits the corruption budget across two mechanisms. This lowers the
intensity of each one and leaves more usable information. Among the pure types Flip
is slightly easier than Loss. Byte-Loss deletes bytes and shifts every following
byte. This is more destructive than the in-place value changes of Bit-Flip.

\subsubsection{Qualitative Comparison}
We visualize the decoding bottleneck on representative images.
Fig.~\ref{fig:qualitative} compares one method per domain under growing
corruption, namely ConvNeXt-Tiny for the pixel domain, the Transform-Bitstream
Classifier (TBC) for the compressed domain, and BRACE for the bitstream domain.
Each column raises the severity from Clean to Extreme. A cell shows the decoded
image when the stream is still decodable and a failure marker otherwise. Even on
the decodable Clean and Light inputs the two baselines are already unreliable, while
BRACE stays correct with high confidence. As severity grows the stream soon becomes
undecodable, and the pixel and compressed methods lose both the image and the
prediction. BRACE reads the raw bytes throughout and keeps predicting the correct
action, even at the Heavy and Extreme levels where decoding always fails. The
examples are curated for illustration and match the quantitative trend reported
above.

\subsection{Ablation Studies}
\label{subsec:ablation}

\subsubsection{Effect of RBCS Augmentation}
Beyond generating the benchmark, RBCS can also serve as a data augmentation by corrupting
bitstreams during training. The injected bit-flips and byte-losses diversify the byte
patterns the model sees. We call this RBCS augmentation. To isolate its effect, we train
ByteFormer with and without RBCS augmentation on BAR-Stanford40 and compare them in
Fig.~\ref{fig:rbcs_ablation}. Augmentation lowers clean accuracy only slightly, yet it
improves accuracy at every corrupted level. The gains are largest under Medium and Heavy
corruption and stay clear at the Extreme level. The same trend holds on BAR-PPMI, shown
in Fig.~10 of the supplementary material. Diversifying the training bitstreams with RBCS
therefore brings a clear robustness gain.

\begin{table}[!t]
    \centering
    \caption{Ablation on the training objectives of Intact-Anchored Representation Alignment (IARA).   
    UAS denotes Unreliable-Anchor Suppression under $\gamma=0.5$. All variants are trained with RBCS augmentation. Acc and mAP report Accuracy and mean Average Precision. Corrupt Avg is the result averaged over all corruption scenarios. 
    These results validate the effectiveness of IARA components.
    }
    \label{tab:recovery_ablation}
    
    \renewcommand{\arraystretch}{1.6} 
    
    \resizebox{\columnwidth}{!}{%
    
    \setlength{\tabcolsep}{3pt}
    
    \begin{tabular}{cccc|cc|cc|cc|cc}
    \toprule
    \multicolumn{4}{c|}{Components} & \multicolumn{4}{c|}{BAR-Stanford40} & \multicolumn{4}{c}{BAR-PPMI} \\
    \cmidrule(lr){1-4} \cmidrule(lr){5-8} \cmidrule(lr){9-12}
    \multirow{2}{*}{$\mathcal{L}_{ce}$} & \multirow{2}{*}{$\mathcal{L}_{feat}$} & \multirow{2}{*}{$\mathcal{L}_{pred}$} & \multirow{2}{*}{UAS} & \multicolumn{2}{c|}{Clean} & \multicolumn{2}{c|}{Corrupt Avg} & \multicolumn{2}{c|}{Clean} & \multicolumn{2}{c}{Corrupt Avg} \\
    & & & & Acc & mAP & Acc & mAP & Acc & mAP & Acc & mAP \\
    \hline
    $\checkmark$ &            &            &            & 60.59 & 63.15 & 29.86 & 29.70 & 48.83 & 50.80 & 21.59 & 22.10 \\
    $\checkmark$ & $\checkmark$ &            &            & 61.98 & 65.83 & 35.40 & 35.68 & 50.06 & 52.11 & 25.07 & 26.18 \\
    $\checkmark$ &            & $\checkmark$ &            & 61.61 & 63.02 & 35.68 & 36.06 & 50.12 & 52.04 & 24.26 & 25.25 \\
    $\checkmark$ & $\checkmark$ & $\checkmark$ &            & 62.02 & 65.97 & 37.23 & 37.65 & 50.36 & 52.31 & 25.80 & 26.94 \\
    $\checkmark$ & $\checkmark$ & $\checkmark$ & $\checkmark$ & \textbf{62.95} & \textbf{66.55} & \textbf{38.75} & \textbf{38.90} & \textbf{51.02} & \textbf{53.01} & \textbf{26.61} & \textbf{27.52} \\
    \bottomrule
    \end{tabular}%
    }
\end{table}

\subsubsection{Effect of IARA Training Objective}

We ablate Intact-Anchored Representation Alignment (IARA) training objective to isolate the contribution of each component. We start from an
RBCS-augmentation baseline that trains the corrupted branch with
$\mathcal{L}_{ce}$ alone. We then add the alignment terms one at a time and report
the results in Table~\ref{tab:recovery_ablation}. Adding either $\mathcal{L}_{feat}$ or $\mathcal{L}_{pred}$ alone improves substantially over the baseline on both subsets. The two terms act at complementary levels. $\mathcal{L}_{pred}$ aligns the predicted distribution of the corrupted
branch with the intact anchor and encourages consistent decisions.
$\mathcal{L}_{feat}$ pulls the corrupted feature toward the intact feature in the
embedding space. Combining them gives a further gain over either one alone, which
confirms that the two levels are complementary. UAS ($\gamma=0.5$) adds the final improvement. An unreliable anchor carries noisy
features and predictions when the intact branch is itself uncertain. Discarding
these anchors makes the alignment signal cleaner. Clean accuracy also rises as we add components. The full model reaches $62.95\%$ on
BAR-Stanford40 and $51.02\%$ on BAR-PPMI. The alignment therefore does not trade
clean accuracy for robustness. Aligning to the intact anchor acts as a regularizer
that improves both.

\section{Conclusion}
To solve the dependency on image decoding and the vulnerability to bitstream corruption in conventional action recognition, this paper proposes a novel Bitstream Action Recognition (BAR) framework \textbf{BRACE}. BRACE employs a dual-branch byte-modeling architecture to generate rich and stable representations of corrupted bitstream. To achieve this, we introduce the Real-world Bitstream Corruption Simulator (RBCS) that reproduces real-world bitstream errors, and build a large-scale BAR dataset (BAR-D) with 13 scenarios and 2 subsets by leveraging RBCS.
A large benchmark is constructed on BAR-D, involving 14 action recognition methods from the pixel, compressed, and bitstream domains.
Extensive experiments demonstrate that BRACE has superior robustness to bitstream corruption than all comparison methods. Ablation studies further validate the effectiveness of the proposed RBCS augmentation and IARA.
This paper provides an alternative path for understanding the underlying semantics of corrupted image bitstream without relying on correcting errors or recovering bitstream. It has the potential to overcome the decoding dependency and corruption vulnerability in real-world storage and transmission scenarios.

\bibliographystyle{IEEEtran}
\bibliography{refs}

\vfill

\end{document}


\title{Bitstream Action Recognition is Byte Modeling}

\title{Supplementary Material for\\
``Bitstream Action Recognition is Byte Modeling''}
\author{
  Fangcheng~Li, 
  Chaoran~Huang,
  Tianyi~Liu,  
  Wenyang~Liu,
  Kejun~Wu,~\IEEEmembership{Senior Member,~IEEE}, 
  Qiong~Liu,~\IEEEmembership{Senior Member,~IEEE},
  You~Yang,~\IEEEmembership{Senior Member,~IEEE},
  and Zhengguo~Li,~\IEEEmembership{Fellow,~IEEE}
  \thanks{Fangcheng~Li, Chaoran~Huang, Kejun~Wu, Qiong~Liu, and You~Yang are with the School of Electronic Information and Communications, Huazhong University of Science and Technology, Wuhan 430074, China.}
  \thanks{Tianyi~Liu and Wenyang~Liu are with School of Electrical and Electronics Engineering, Nanyang Technological University, Singapore.}
  \thanks{Zhengguo~Li is with VI Department, Institute for Infocomm Research, Agency for Science, Technology and Research (A*STAR), Singapore.}  
  \thanks{This work was supported by the National Natural Science Foundation of China under Grant 62501246.}
  \thanks{Corresponding author: Kejun~Wu (kjwu@hust.edu.cn).}
  }

\maketitle

\setcounter{figure}{8}  

The main text presents all analyses on BAR-Stanford40, while this supplementary
material provides the corresponding results on BAR-PPMI. The same protocol and figure
layout are used, so each supplementary figure can be directly compared with its
counterpart in the main text. BAR-PPMI is the smaller subset and has a different action
vocabulary, mainly involving people interacting with musical instruments. Consistent
findings across the two subsets therefore better reflect the task itself rather than
dataset-specific properties.

\begin{figure}[!t]
    \centering
    \begin{subfigure}{0.62\linewidth}
        \includegraphics[width=\linewidth]{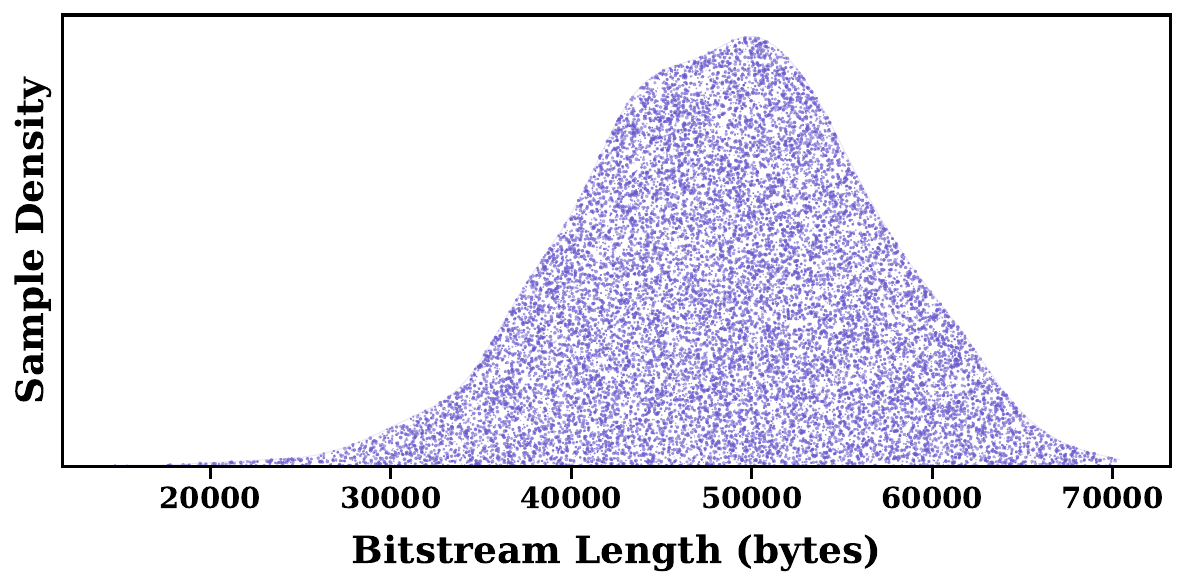}
        \caption{Bitstream Distribution}
    \end{subfigure}

    \begin{subfigure}{0.62\linewidth}
        \includegraphics[width=\linewidth]{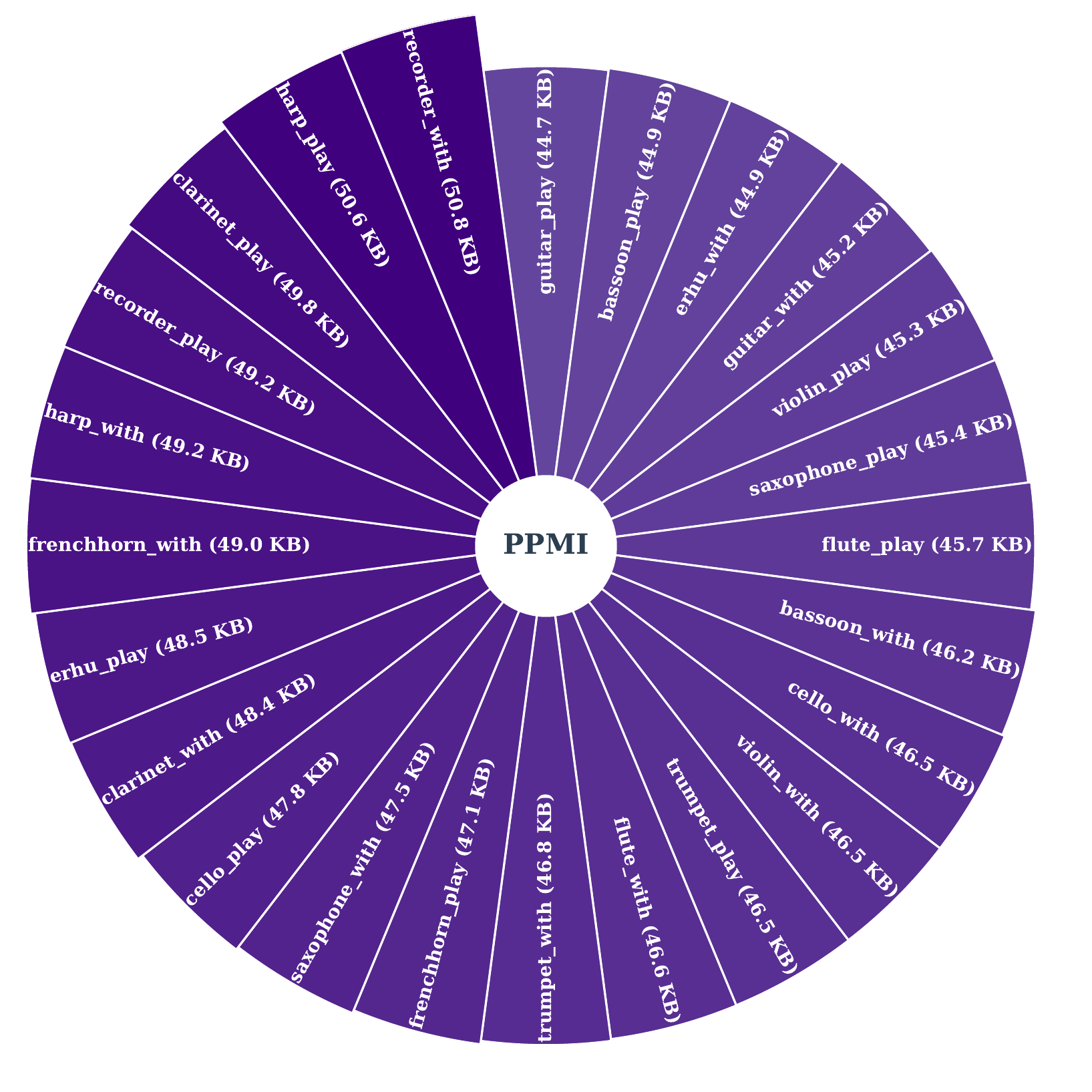}
        \caption{Average Length by Class}
    \end{subfigure}

    \caption{Bitstream length statistics on BAR-PPMI. (a) Distribution of bitstream lengths
over all samples. (b) Average bitstream length per action class. The patterns match those
on BAR-Stanford40 in the main text, and length offers no shortcut for recognition.}
    \label{fig:bitstream_analysis_ppmi}
\end{figure}

Figure~\ref{fig:bitstream_analysis_ppmi} repeats the bitstream length analysis. The top
panel shows that the sample lengths concentrate in a narrow band, without a clear
secondary mode that could provide an obvious shortcut. The bottom panel shows that the
per-class average lengths remain close to one another, so no action class is clearly
separated by file size. This suggests that bitstream length does not provide a reliable
shortcut for recognition on BAR-PPMI, consistent with the observation on BAR-Stanford40.

\begin{figure}[!t]
  \centering
  \includegraphics[width=\linewidth]{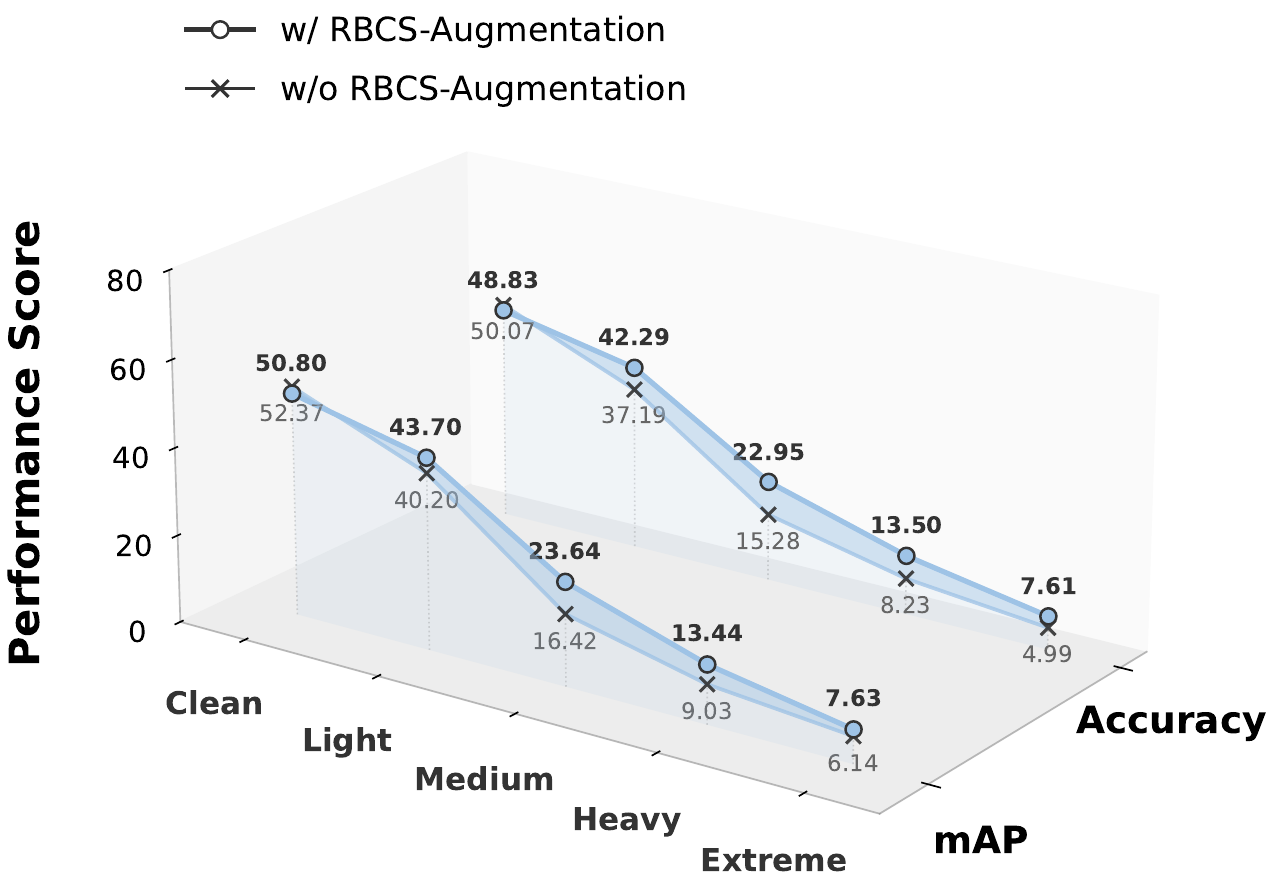}
  \caption{Ablation on the effectiveness of RBCS augmentation on BAR-PPMI.}
  \label{fig:rbcs_ablation_ppmi}
\end{figure}

Figure~\ref{fig:rbcs_ablation_ppmi} repeats the RBCS augmentation ablation. RBCS
augmentation only slightly reduces clean accuracy, while improving performance at every
corrupted level. The gains are most pronounced in the middle severity range and remain
clear at the Extreme level, showing the same trend as on BAR-Stanford40. Diversifying
the training bitstreams with RBCS therefore brings a consistent robustness gain on
BAR-PPMI as well.

\begin{figure*}[!htbp]
  \centering
  \includegraphics[width=\linewidth]{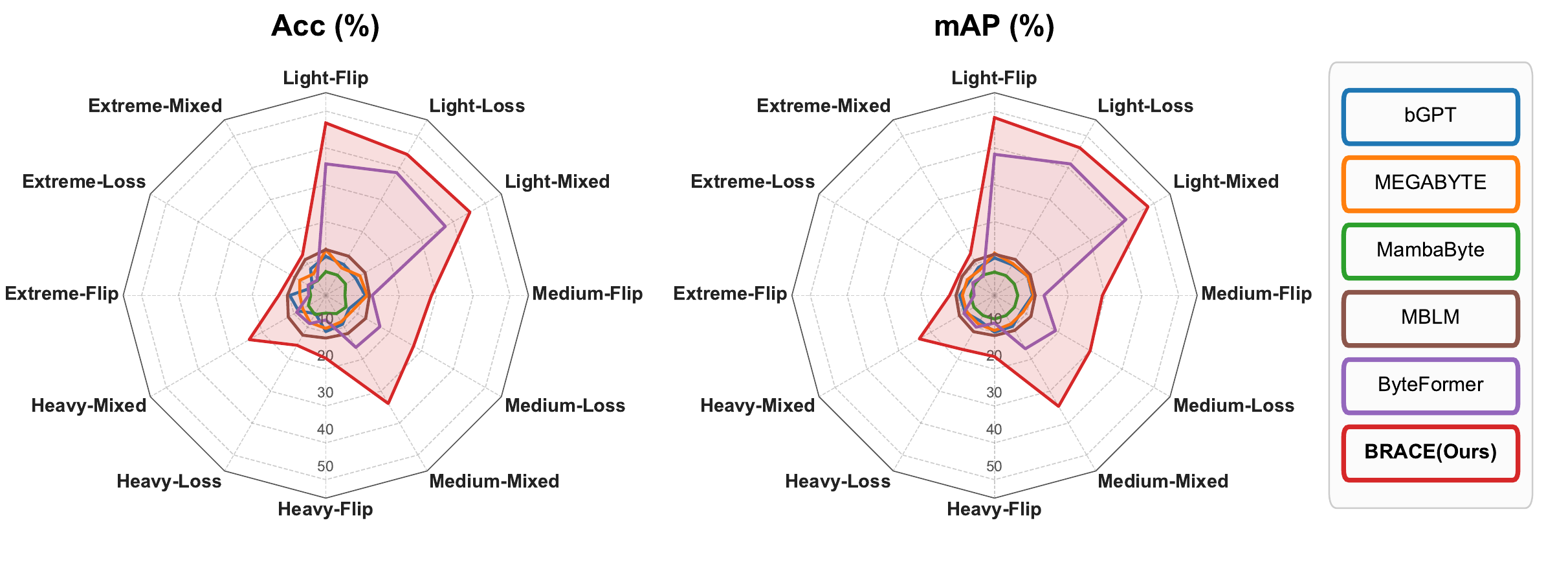}
  \caption{Per-scenario comparison of bitstream-domain methods on BAR-PPMI in terms of Accuracy (\%) and mAP (\%). Each axis represents one of the 12 corruption scenarios. Our BRACE outperforms all comparison methods across all scenarios.}
  \label{fig:12_scenario_ppmi}
\end{figure*}

Figure~\ref{fig:12_scenario_ppmi} compares the bitstream-domain methods across all 12
corruption scenarios, with Top-1 Accuracy on the left and mAP on the right. BRACE traces
the outermost contour in both panels, indicating that it performs best across the
corruption scenarios rather than only on average. As severity increases from Light to
Extreme, the contours of all methods contract toward the center, but BRACE keeps the
widest envelope throughout. ByteFormer remains the closest competitor, which is
consistent with the ordering observed on BAR-Stanford40.

\begin{figure*}[!htbp]
  \centering
  \includegraphics[width=1\linewidth]{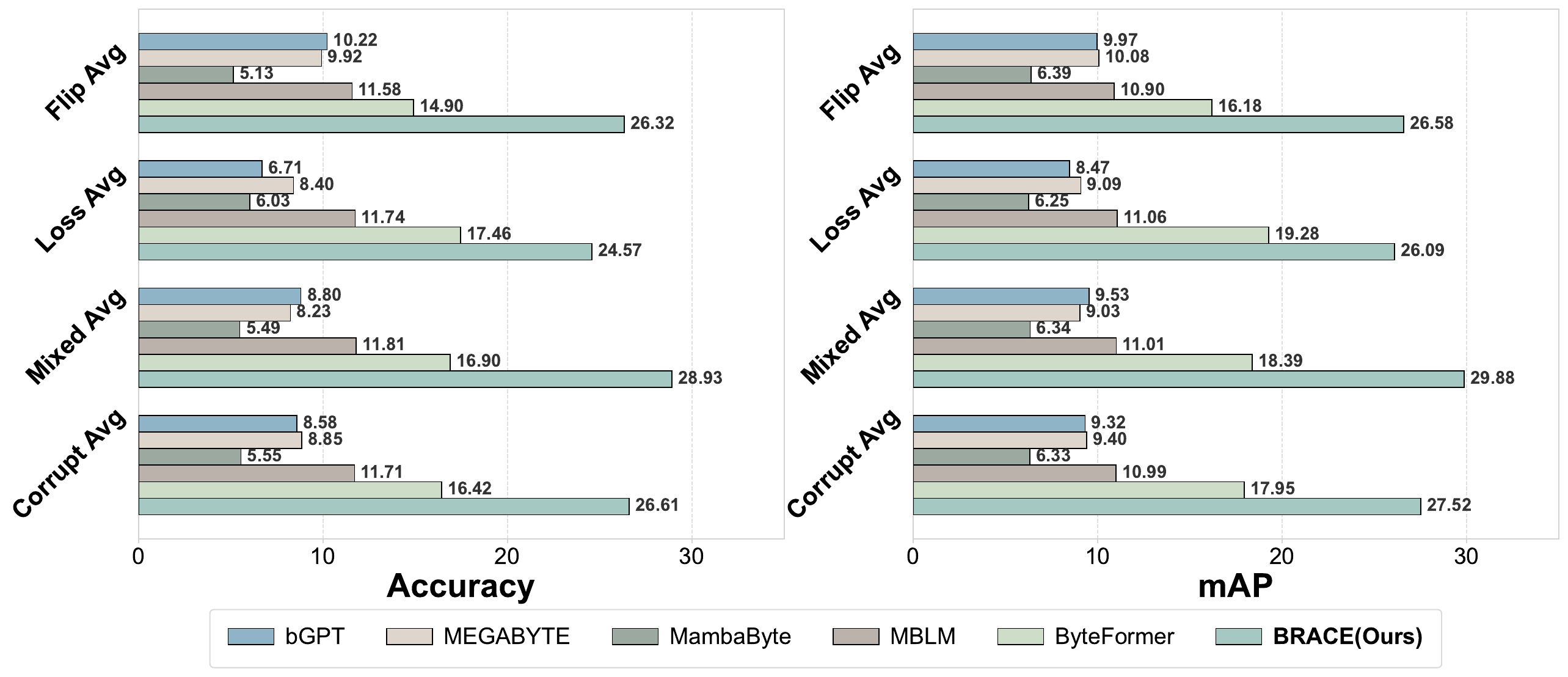}
  \caption{Comparison of bitstream-domain methods by corruption type on BAR-PPMI. The results report the average Accuracy and mAP for four severity levels on different corruption types. Our BRACE outperforms all comparison methods across all corruption types.}
  \label{fig:Different_Corrupt_type_ppmi}
\end{figure*}

Figure~\ref{fig:Different_Corrupt_type_ppmi} averages the four severity levels within
each corruption type. BRACE achieves the best Accuracy and mAP under Flip, Loss, and
Mixed, demonstrating its consistent advantage across different corruption mechanisms.
The results further confirm that the robustness gain of BRACE on BAR-PPMI is not limited to a particular corruption type.

Overall, the BAR-PPMI results align well with those on BAR-Stanford40 across bitstream length analysis, per-scenario robustness, corruption-type comparison, and RBCS augmentation. These consistent findings suggest that the main observations are not specific to one subset, but hold across datasets with different scales and action vocabularies.
